\documentclass[conference]{IEEEtran}
\IEEEoverridecommandlockouts

\usepackage{cite}
\usepackage{amsmath,amssymb,amsfonts}
\usepackage{algorithmic}
\usepackage{graphicx}
\usepackage{textcomp}

\usepackage[inline]{enumitem}
\usepackage{algorithmic}
\usepackage{graphicx}
\usepackage{textcomp}
\usepackage{xcolor}

\usepackage{witharrows}

\usepackage[flushleft]{threeparttable}
\usepackage{tablefootnote}
\usepackage{array}

\newcolumntype{P}[1]{>{\centering\arraybackslash}m{#1}}
\usepackage{multicol}
\usepackage{multirow}
\usepackage{tabulary}
\usepackage{colortbl}
\definecolor{mygray}{gray}{0.90}
\usepackage{makecell}
\usepackage{booktabs}
\usepackage{subcaption}
\usepackage{stmaryrd}
\usepackage{float}
\usepackage{pifont}

\usepackage{arydshln}
\usepackage{array}
\newcolumntype{P}[1]{>{\centering\arraybackslash}p{#1}}

\colorlet{mygray}{gray!15!white}

\usepackage{url,hyperref,microtype}
\hypersetup{
    colorlinks=true,
    citecolor=blue,
    linkcolor=blue,
    filecolor=magenta,      
    urlcolor=black,
}

\def\BibTeX{{\rm B\kern-.05em{\sc i\kern-.025em b}\kern-.08em
T\kern-.1667em\lower.7ex\hbox{E}\kern-.125emX}}
\begin{document}

\title{Efficient Emotion-Aware Iconic Gesture Prediction for Robot Co-Speech
}

\author{\IEEEauthorblockN{Edwin C. Montiel-Vazquez}
\IEEEauthorblockA{\textit{School of Engineering and Sciences} \\
\textit{Tecnologico de Monterrey} 
State of Mexico, Mexico  \\
edwincmv@exatec.tec.mx}
\and
\IEEEauthorblockN{Christian Arzate Cruz}
\IEEEauthorblockA{\textit{Honda Research Institute Japan} \\
Wako City, Japan  \\
christian.arzate@jp.honda-ri.com}
\and
\IEEEauthorblockN{Stefanos Gkikas}
\IEEEauthorblockA{\textit{Honda Research Institute Japan} \\
Wako City, Japan \\
stefanos.gkikas@jp.honda-ri.com}
\and
\IEEEauthorblockN{Thomas Kassiotis}
\IEEEauthorblockA{\textit{Department of Electronic Engineering} \\
\textit{Hellenic Mediterranean University}\\
Chania, Greece \\
ddk305@edu.hmu.gr}
\and
\IEEEauthorblockN{Giorgos Giannakakis}
\IEEEauthorblockA{\textit{Department of Electronic Engineering} \\
\textit{Hellenic Mediterranean University}\\
Chania, Greece \\
ggian@hmu.gr}
\and
\IEEEauthorblockN{Randy Gomez}
\IEEEauthorblockA{\textit{Honda Research Institute Japan} \\
Wako City, Japan \\
r.gomez@jp.honda-ri.com}
}

\maketitle

\begin{abstract}
Co-speech gestures increase engagement and improve speech understanding. Most data-driven robot systems generate rhythmic beat-like motion, yet few integrate semantic emphasis. To address this, we propose a lightweight transformer that derives iconic gesture placement and intensity from text and emotion alone, requiring no audio input at inference time. The model outperforms GPT-4o in both semantic gesture placement classification and intensity regression on the BEAT2 dataset, while remaining computationally compact and suitable for real-time deployment on embodied agents.
\end{abstract}


\section{Introduction}

Emotional expressiveness is fundamental to natural and engaging communication. Humans convey their internal state through body gestures and facial expressions, ranging from large deliberate movements that depict the meaning of what is said, known as \emph{iconic} or \emph{semantic} gestures, to small rhythmic motions that follow speech rhythm, known as \emph{beat} gestures~\cite{nyatsanga2023comprehensive}. Prior work on robot co-speech gesture generation has largely focused on beat gestures, while semantic gestures remain rarely addressed, despite growing interest in the field~\cite{liu2024emage, zhang2025semtalk}.

A further limitation of existing methods is that they do not explicitly model how emotion shapes movement~\cite{neff2008gesture, kucherenko2020gesticulator, wu2021probabilistic}. The most related work is by Ishii \textit{et al}. \cite{ishii2025impact}, who condition pose generation on personality traits. However, emotion — not personality — is what most directly drives physical expression. We therefore condition our system on four basic emotions from Plutchik's wheel \cite{Plutchik2001nature}: joy, anger, sadness, and fear. 
The importance of affective modeling in text-driven communication systems extends beyond robotics; even frameworks designed specifically for analyzing human language in social platforms have identified the absence of emotion recognition as a central limitation, highlighting the broader need for emotion-aware approaches to text processing \cite{chatziadam_dimitraidis_2020}.
Given an utterance and a target emotion, our model produces a gesture sequence that integrates both beat and iconic motion, enabling a robot to express not only \emph{what} it says but also \emph{how} it feels, in real time.

\IEEEpubidadjcol

Most co-speech methods for robots~\cite{kucherenko2020gesticulator,yoon2019robots,li2024learning,fernandez2025evaluating} and artificial agents~\cite{wu2021probabilistic, neff2008gesture, ishii2025impact, omine2025co} assume audio is available at inference time to extract prosodic features or synchronize motion. For robots that rely on text-to-speech (TTS), this assumption introduces latency and reduces responsiveness. While LLMs can integrate semantic context effectively, their computational cost makes them impractical for real-time deployment on most robots and embodied agents. We therefore propose a text-only, emotion-aware pipeline that takes two inputs: the utterance the robot will say and the intended emotion. We build on the \textbf{b}ody-\textbf{e}xpression-\textbf{a}udio-\textbf{t}ext (BEAT2) dataset~\cite{liu2024emage} to train a model that identifies semantically relevant words in an utterance and quantifies their gesture intensity~\cite{zhang2025semtalk}. 

\begin{figure}
\begin{center}
\includegraphics[scale=0.14]{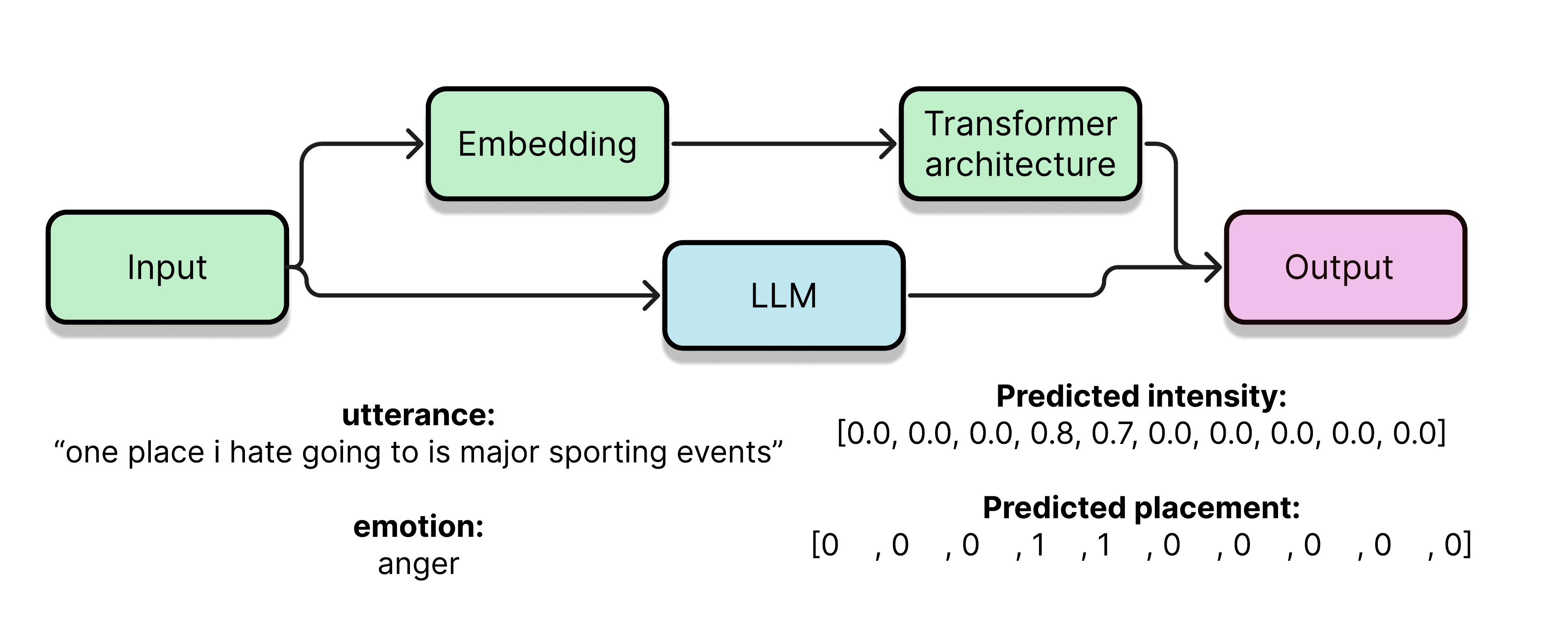} 
\end{center}
\caption{Task overview. An utterance is separated into words, and the semantic gesture placement and intensity is calculated per word.}
\label{fig:overview}
\end{figure}

Our approach is presented in Figure~\ref{fig:overview}. The contributions of this work are: (i) a text-based model for semantic gesture placement in sentences, (ii) an efficient method for iconic gesture intensity regression, and (iii) a framework for emotion-aware semantic gestures in social robots.

\section{Related Work} \label{sec:related_work}
In this section, we review prior work on co-speech gesture generation for artificial agents and robots.

\subsection{Artificial Agents}

Co-speech gesture generation for artificial agents has received considerable attention, particularly in recent years. Among the most important contributions to the field are the BEAT~\cite{liu2022beat} and BEAT2~\cite{liu2024emage} datasets. Based on monologues, both combined provide over $70$ hours of motion-capture recordings and mesh representations paired with audio, text, and frame-level semantic labels. Of particular relevance, BEAT2 includes manually annotated iconic gesture intensity values at the word level. We use BEAT2, the extension of BEAT, to train our iconic gesture placement and intensity models.

Recent work has shifted toward full-body motion generation~\cite{gao2025sarges}, moving away from earlier upper-body approaches~\cite{wu2021probabilistic, neff2008gesture, ishii2025impact}. Generating accurate, human-like motion requires accounting for variability in emotion and semantic emphasis, two aspects that remain underexplored. Computational efficiency is an additional requirement for real-time robotics applications.

Model architectures have evolved from recurrent networks such as long short-term memory (LSTM)~\cite{Hochreiter1997long} to attention-based transformers~\cite{vaswani2017attention}. Adversarial and diffusion-based methods have also been proposed to improve motion realism and diversity~\cite{liu2022beat, liu2024emage, yang2023diffusestylegesture}. We adopt a transformer-based architecture, as the attention mechanism is well-suited for text-driven tasks while remaining computationally efficient.

Applications range from realistic gesture synthesis~\cite{zhang2025semtalk} to stylized 3D character animation~\cite{omine2025co}, showing that models trained on human motion data can generalize across different embodiments. This supports the use of a single model trained on human data for social robot applications.

\subsection{Robots}

Embodied co-speech gesture generation has been explored less thoroughly than its counterpart for artificial agents. Approaches for semantic gestures in robots are sparse, ranging from re-targeting human motion~\cite{go2018andorid} to rule-based policies~\cite{bremner2009beat}. Most data-driven methods focus on rhythmic motion, using audio and text to generate head and arm movements that follow speech rhythm. The main differences between methods lie in how linguistic and prosodic information is represented and in the learning architectures used.

For linguistic features, sentence-level context has been addressed using BERT embeddings~\cite{kucherenko2020gesticulator, devlin2019bert}, while word-level conditioning enables finer lexical control~\cite{yoon2019robots}. Gestures are commonly modeled as a regression problem over joint poses using autoregressive architectures~\cite{kucherenko2020gesticulator, yoon2019robots, li2024learning, fernandez2025evaluating}, which generate each new joint position from previous poses to maintain motion continuity. GANs have also been used to better match the distribution of human motion while reducing over-smoothing~\cite{liu2023speech, yu2020srg}.

A key limitation is that the few robotics methods targeting iconic gestures typically trigger pre-defined animations without modeling placement timing or intensity~\cite{fernandez2025evaluating}. In contrast, our model explicitly predicts where iconic gestures should occur and how strongly they should be performed.

\section{Proposed Approach} \label{sec:system_overview}

As presented in Figure~\ref{fig:overview}, our semantic co-speech pipeline encodes the input through an embedding layer and passes it to a transformer model. The embeddings are drawn from established language models~\cite{devlin2019bert, reimers2019sentence} and emotion representations~\cite{xu2018emo2vec}. At inference, the system takes a text prompt and a target emotion as input and outputs a time-aligned sequence specifying the placement and intensity of semantic-emphasis gestures.

The text is encoded with SBERT~\cite{reimers2019sentence}, while word-level embeddings are obtained using emo2vec~\cite{xu2018emo2vec}. The word embeddings are combined with the utterance's emotion by averaging the word and emotion-label representations. The model then predicts iconic-gesture placement at the word level and the corresponding intensity, conditioned on the target emotion, thereby reflecting the affective characteristics of the utterance. Training details are described in the following section.

\section{Iconic Gesture Prediction} \label{sec:iconics}

\subsection{Inputs and Supervision}

The two inputs to our system are the text prompt the robot will say and the emotion of the utterance. Both are obtained from the BEAT2 dataset~\cite{liu2024emage} for training. BEAT2 provides word-level iconic annotations as continuous intensity values, which we use as regression targets for intensity and classification targets for placement. The intensity labels are binarized by setting the activation to $1$ when the value exceeds $0.5$ and to $0$ otherwise, following the threshold used in previous work~\cite{zhang2025semtalk}.

The data is organized into sentence and word-of-interest pairs $p_n = (s, w_n)$, where $s = (w_1, w_2, \ldots, w_{40})$ represents the full sentence and $n$ identifies the word of interest within $s$. We encode $s$ using SBERT~\cite{reimers2019sentence} to obtain sentence-level semantic embeddings $h_s \in \mathbb{R}^{384}$, shared across all words $w_n$ in the sentence. For word-level representation, we use emo2vec~\cite{xu2018emo2vec} to obtain $e_w = \text{emo2vec}(w_n)$, where $e_w \in \mathbb{R}^{100}$. To incorporate the overall sentence emotion, $e_w$ is augmented with $e_{emo} = \text{emo2vec}(\text{label})$, where $\text{label}$ is the emotion label of the sentence. The emotion-enhanced word representation is then $e_n = (e_w + e_{emo})/2$. The final input used to predict the iconic label $c$ or intensity $i$ for each word is $p_n = (h_s, e_n)$.

\subsection{Transformer Architecture}

The proposed architecture leverages cross- and self-attention to enable efficient global modeling with reduced computational complexity \cite{gkikas_kyprakis_resp_2025, gkikas_tsiknakis_thermal_2024, gkikas_arzate_eeite_pain_2026, gkikas_cruz_eeite_cwl_2026}. 
Rather than applying attention directly to all input embeddings, a compact latent space is introduced as an intermediate representation. The flattened input embeddings are represented as $\mathbf{X} \in \mathbb{R}^{M \times D}$, while a learnable latent matrix $\mathbf{Z}_0 \in \mathbb{R}^{N \times d}$ with $N \ll M$ aggregates information from the input, forming an efficient bottleneck.
Positional information is incorporated using Fourier feature encoding:
\begin{equation}
\gamma(p) = [\sin(\pi \omega_k p), \cos(\pi \omega_k p), p]_{k=1}^{K},
\end{equation}
where $p$ denotes a normalized coordinate and $\omega_k$ the frequency bands.
The encoded features are concatenated with the input embeddings before attention. Cross-attention maps the input into the latent space:
\begin{equation}
\mathrm{Attn}_{\text{cross}}(\mathbf{Z}, \mathbf{X}) =
\mathrm{softmax}\!\left(\frac{\mathbf{Q}_Z \mathbf{K}_X^\top}{\sqrt{d_h}}\right)\mathbf{V}_X ,
\end{equation}
with $\mathbf{Q}_Z$, $\mathbf{K}_X$, and $\mathbf{V}_X$ denoting query, key, and value projections.
The latent representations are then processed by self-attention within the latent space, enabling global interactions among latent tokens. Each attention block is followed by a feedforward transformation:
\begin{equation}
\mathrm{FFN}(\mathbf{z}) = W_2\,\mathrm{GELU}(W_1 \mathbf{z}),
\end{equation}
where $W_1$ and $W_2$ are learned parameters.
Finally, the latent embeddings are mean-pooled and projected through a fully connected layer for prediction. The implemented configuration uses $128$ latent tokens of dimension $256$, $1$ cross-attention head, and $8$ self-attention heads.

\begin{figure}
\begin{center}
\includegraphics[scale=0.12]{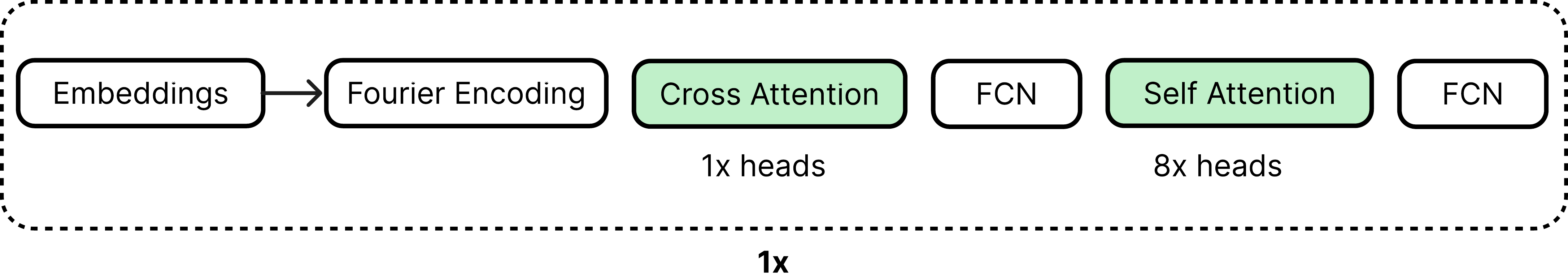} 
\end{center}
\caption{High-level overview of the proposed model.}
\label{fig:model}
\end{figure}

\section{Experimental Design}


\subsection{Dataset Preparation}

For training and evaluation, we use the BEAT2 dataset~\cite{zhang2025semtalk, liu2024emage}, which represents the state of the art for full-body gesture tracking. Despite its extensive use for gesture generation, the semantic emphasis and iconic gesture aspects of the dataset remain underexplored.

BEAT2 contains around 2,000 entries of variable length, which, given their nature as monologue speeches, also vary in the number of iconic gestures present. To improve training and allow the model to better handle sparse iconic activations, we segment the data into sentences, yielding approximately 18,000 data points. Each data point consists of an utterance paired with one of the following emotions: sadness, neutral, anger, contempt, surprise, disgust, fear, or happiness. For consistency with Plutchik's emotion model~\cite{Plutchik2001nature}, validated across psychology and affective computing~\cite{montiel-vazquez_explainable_2022, Montiel2024}, we rename the label `happiness' to `joy'. All results are reported on the test set using an 80/20 train-test split.

\subsection{Baseline}

We compare our model against GPT-4o~\cite{openai_gpt-4_2024} as a baseline, given the demonstrated capability of LLMs in understanding semantic information~\cite{havlik2024meaning}. The model was prompted to predict the intensity of iconic gestures per word, in the same format as the dataset, conditioned on the utterance's emotion. The resulting values were binarized using the same threshold as our model to obtain placement predictions.

\subsection{Model Configurations}

Since computational efficiency is a primary objective, we evaluate two architectural parameters: the number of cross-attention layers (Depth) and the number of sequential self-attention (SA) blocks. We explore the minimum configuration that delivers strong performance.

\section{Experimental Results}

\subsection{Model Size}
Table~\ref{tab:model_size} shows the results across all configurations. Classification performance remains stable, with accuracy ranging between $68.53\%$ and $68.78\%$, indicating that the task does not benefit from additional capacity. Computational cost, however, varies considerably: GFLOPs drop from $5.79$ to $0.55$ and latency from $8.39$ ms to $1.16$ ms as the model shrinks. We therefore select depth 1 with a single SA block for all subsequent experiments, as it achieves competitive performance at the lowest computational cost. This confirms that a minimal architecture is sufficient for this task, which is a desirable property for real-time robot deployment.

\begin{table}
\centering
\caption{Performance and computational cost across model configurations.}
\label{tab:model_size}
\footnotesize
\begin{threeparttable}
\begin{tabular}{P{0.5cm} P{0.3cm} P{0.9cm} P{0.9cm} P{0.5cm} P{0.9cm} P{1.8cm}}
\toprule
\textbf{Depth} & \textbf{SA} & \textbf{Accuracy} & \textbf{Precision} & \textbf{F1} & \textbf{GFLOPs} & \textbf{Latency (ms)$\downarrow$} \\
\midrule
2 & 8 & 68.78 & 53.92 & 50.27 & 5.79 & 8.39 \\
2 & 4 & 68.64 & 53.55 & 47.84 & 3.11 & 4.46 \\
2 & 2 & 68.75 & 53.82 & 49.72 & 1.77 & 3.20 \\
2 & 1 & 68.68 & 53.76 & 49.38 & 1.09 & 2.16 \\
1 & 8 & 68.53 & 53.76 & 49.57 & 2.90 & 4.02 \\
1 & 4 & 68.56 & 53.68 & 48.98 & 1.55 & 2.45 \\
1 & 2 & 68.59 & 53.47 & 47.59 & 0.78 & 1.74 \\
\rowcolor{mygray}
1 & 1 & 68.64 & 53.55 & 47.84 & 0.55 & 1.16 \\
\bottomrule
\end{tabular}
\begin{tablenotes}
\scriptsize
\item SA: self-attention blocks. The selected configuration is highlighted.
\end{tablenotes}
\end{threeparttable}
\end{table}

\subsection{Iconic Placement}

Table~\ref{tab:iconic_macro_results_cls} reports macro-averaged results for iconic gesture placement. Our model outperforms the LLM baseline across all metrics, with a notable improvement in accuracy ($68.64\%$ vs. $53.36\%$). The lower F1 score relative to accuracy reflects the class imbalance inherent in the task, as iconic gestures are sparse within utterances, making precise per-word prediction challenging for both models.

\begin{table}
\centering
\caption{Macro-averaged results for iconic placement per word.}
\label{tab:iconic_macro_results_cls}
\footnotesize
\begin{threeparttable}
\begin{tabular}{lcccc}
\toprule
\textbf{Model} & \textbf{Accuracy} & \textbf{Precision} & \textbf{Recall} & \textbf{F1} \\
\midrule
\textbf{LLM}  & 53.36 & 52.63 & 53.36 & 52.92 \\
\textbf{Ours} & \underline{68.64} & \underline{53.55} & \underline{68.64} & \underline{47.84} \\
\bottomrule
\end{tabular}
\begin{tablenotes}
\scriptsize
\item F1 = F1 score. Underlined values are best per column.
\end{tablenotes}
\end{threeparttable}
\end{table}

\begin{table}
\centering
\caption{Regression results for word intensity of iconic gestures.}
\label{tab:iconic_macro_results}
\footnotesize
\begin{threeparttable}
  \begin{tabular}{lcccccc}
    \toprule
    \textbf{Model} & \textbf{MAE} & \textbf{MSE} & \textbf{RMSE} & \textbf{R\textsuperscript{2}} & \textbf{PR} & \textbf{Spearman} \\
    \midrule
    \textbf{LLM}  & 0.08 & 0.05 & 0.22 & -1.23 & 0.09 & 0.06 \\
    \textbf{Ours} & 0.08 & 0.02 & 0.15 & -0.07 & 0.20 & 0.16 \\
    \bottomrule
  \end{tabular}
  \begin{tablenotes}
    \scriptsize
    \item MAE = Mean Absolute Error, MSE = Mean Squared Error, RMSE = Root Mean Squared Error,
    R\textsuperscript{2} = R-squared (coefficient of determination), PR = Pearson's correlation coefficient,
    Spearman = Spearman's rank correlation coefficient.
  \end{tablenotes}
\end{threeparttable}
\end{table}

\begin{figure*}[ht]
\begin{center}
\includegraphics[scale=0.15]{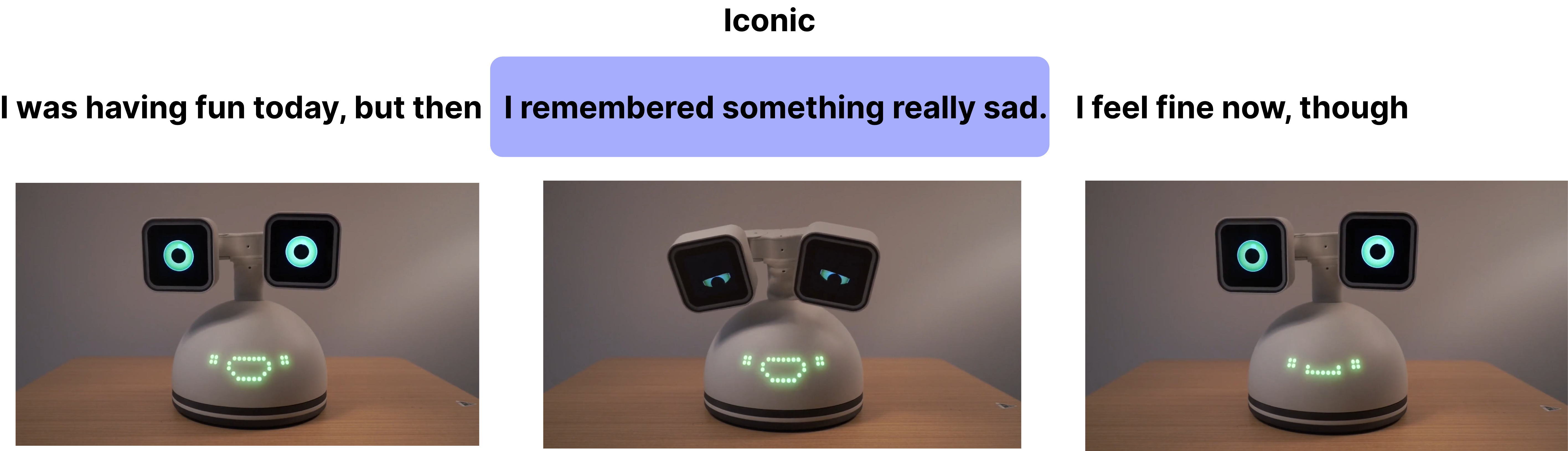} 
\end{center}
\caption{Semantic co-speech implementation on the social robot Haru.}
\label{fig:implementation}
\end{figure*}

\subsection{Intensity Regression}

Table~\ref{tab:iconic_macro_results} reports regression results for iconic gesture intensity. Despite the task being challenging, our model outperforms GPT-4o across all metrics, improving RMSE from $0.22$ to $0.15$ and Pearson correlation from $0.09$ to $0.20$. The negative R\textsuperscript{2} values for both models indicate that intensity prediction remains an open problem, likely due to the dataset's subjective and sparse iconic gesture annotations.

\section{Discussion}

The results show that a lightweight, text-only model can outperform GPT-4o on both iconic gesture placement and intensity regression, using only the utterance and the target emotion as input. This suggests that task-specific training on word-level iconic annotations provides a stronger inductive bias for this problem than the general semantic knowledge encoded in large pretrained models.

Placement results are strong, with our model achieving $68.64\%$ accuracy against $53.36\%$ for the LLM baseline. Intensity regression remains more challenging for both models, as reflected by the negative R\textsuperscript{2} values. This is likely due to the dataset's subjective and sparse iconic gesture annotations, as well as the limited expressiveness of the current word-level embeddings. Exploring richer semantic representations is a natural direction for future work, alongside larger and more diverse datasets for this task, which remains underexplored in affective computing for embodied agents.

\section{Robot Implementation}

Our model can be deployed alongside any co-speech approach that handles rhythmic motion~\cite{kucherenko2020gesticulator, yoon2019robots, li2024learning, liu2023speech, yu2020srg}, as it operates independently by predicting the placement and intensity of iconic gestures from text. We deploy it on the social robot Haru~\cite{Gomez2018haru}, as shown in Figure~\ref{fig:implementation}. Iconic gesture intensity values are mapped to a set of animations corresponding to the detected emotion and intensity level. When the model identifies a word that requires an iconic gesture, the robot executes the corresponding animation in real time.
Although further evaluation across different robot platforms is needed, the implementation demonstrates the feasibility of the proposed approach in a real-world setting.

\section{Conclusions}

We presented a lightweight, emotion-aware transformer for semantic gesture placement and intensity prediction in robot co-speech. Taking only text and a target emotion as input, the model outperforms GPT-4o on both tasks while remaining computationally compact, with a latency of $1.16$ ms on GPU. The implementation on Haru robot demonstrates its applicability to real-time embodied agents. 
The need for low-latency, real-time inference is a recognized challenge across a wide range of application domains~\cite{antonogiorgakis_britzolakis_2019}; our model, with a latency of 1.16~ms, directly addresses this constraint, remaining lightweight enough for deployment on embodied agents in real-world settings~\cite{arzate_3dmm_2022}.
Future work should focus on improving intensity regression using richer embeddings and on generalizing the approach to other robot platforms and co-speech scenarios beyond iconic gestures, including gaze-aware and perceptually-grounded behaviours~\cite{hessels_fang_2026}.

\bibliographystyle{IEEEtran}
\bibliography{library}

\end{document}